\documentclass[conference]{IEEEtran}
\IEEEoverridecommandlockouts

\usepackage{cite}
\usepackage{amsmath,amssymb,amsfonts}
\usepackage{algorithmic}
\usepackage{graphicx}
\usepackage{textcomp}
\usepackage{xcolor}
\usepackage{times} 
\usepackage{amsmath} 
\usepackage{amssymb}  
\usepackage{multirow}
\usepackage{multicol}
\usepackage{url}
\usepackage{afterpage}
\usepackage{stfloats}
\usepackage{cite}
\usepackage{arydshln}
\usepackage{epsfig} 
\usepackage{mathptmx}
\usepackage{subcaption}
\usepackage{tablefootnote}
\usepackage{enumitem}

\def\BibTeX{{\rm B\kern-.05em{\sc i\kern-.025em b}\kern-.08em
    T\kern-.1667em\lower.7ex\hbox{E}\kern-.125emX}}
\begin{document}

\title{MutualForce: Mutual-Aware Enhancement for 4D Radar-LiDAR 3D Object Detection\\
}

\author{
    Xiangyuan Peng$^{\star \dagger}$ \qquad Huawei Sun$^{\star  \dagger}$ \qquad Kay Bierzynski$^{\dagger}$ 
    \qquad Anton Fischbacher$^{\star}$ \\  Lorenzo Servadei$^{\star}$ \qquad Robert Wille$^{\star}$\\
    \\
    $^{\star}$ Technical University of Munich, Munich, Germany \\
    $^{\dagger}$ Infineon Technologies AG, Neubiberg, Germany
}

\maketitle

\begin{abstract}
Radar and LiDAR have been widely used in autonomous driving and robotics as LiDAR provides rich structure information, and radar demonstrates high robustness under adverse weather. Recent studies highlight the effectiveness of fusing radar and LiDAR point clouds. However, challenges remain due to the modality misalignment and information loss during feature extractions.
To address these issues, we propose a 4D radar-LiDAR framework to mutually enhance their representations. 
Initially, the indicative features from radar are utilized to guide both radar and LiDAR geometric feature learning. Subsequently, to mitigate their sparsity gap, the shape information from LiDAR is used to enrich radar BEV features. 
Extensive experiments on the View-of-Delft (VoD) dataset demonstrate our approach's superiority over existing methods, achieving the highest mAP of 71.76\% across the entire area and 86.36\% within the driving corridor. 
Especially for cars, we improve the AP by 4.17\% and 4.20\% due to the strong indicative features and symmetric shapes.  
\end{abstract}

\begin{IEEEkeywords}
3D object detection, Sensor fusion, Indicative feature, Shape information, Autonomous driving.
\end{IEEEkeywords}

\section{Introduction}
\label{sec:intro}
Object detection is an important task in autonomous driving and robotics. Many detection algorithms have been developed based on various sensors, including cameras, LiDAR, and radar \cite{peng2024scene,rollo2023carpe,rothmeier2024time,liu2024revisiting}.

Cameras benefit from advanced 2D detection algorithms but lack depth perception and can raise privacy concerns \cite{10446563}. In contrast, LiDAR point clouds provide detailed 3D geometric structures. Current LiDAR-based detection algorithms can be categorized into voxel-wise, point-wise, and hybrid-wise approaches. Voxel-wise methods partition point clouds into voxels or pillars \cite{chen2023svqnet}, while point-wise methods extract features directly from raw points \cite{liu2024rethinking}. Some algorithms combine two methods to balance the computation cost and information loss \cite{shi2020pv}.
However, the performance of LiDAR detection decreases over longer distances and is not robust in adverse conditions \cite{chae2024towards}. Small particles such as rain, fog, or dust can introduce noise to LiDAR point clouds \cite{hahner2021fog}, leading to false negative detections. 
Besides, LiDAR lacks velocity information, which is crucial for dynamic objects.

Consequently, radar has gained increased attention. Compared to LiDAR, radar remains robust performance over long distances and under extreme weather \cite{paek2022k,Saini_2024_CVPR,zhang2023dual}. Besides, radar can reflect dynamic and material information by velocity and Radar Cross-Section (RCS) measurement. In particular, 4D radar additionally provides height information, contributing to advancing radar-based detection in challenging environments \cite{palffy2022multi,zheng2022tj4dradset,zhang2023dual}.
However, 4D radar point clouds remain sparser compared to those from LiDAR, making it challenging to detect small and low-speed objects. Therefore, the practical application of radar-only detection is still limited.

To overcome the limitation of radar and LiDAR-only methods, a notable research direction is to fuse 4D radar and LiDAR point clouds \cite{wang2022interfusion,roldan2024see,meng2024traffic}, which combines accurate spatial information with high robustness. Some methods combine 4D radar and LiDAR points in the voxels or pillars, then use one joint encoder to extract united features \cite{song2024lirafusion,wang2023bi}. Other methods utilize two backbones to extract features parallelly and fuse them at the bird's-eye view (BEV) stage \cite{yang2022ralibev,9578621,wang2023bi}. 
However, most existing 4D radar-LiDAR fusion methods have not fully leveraged the unique strengths of each sensor to effectively interact with each other. The shortcomings can lead to information loss and misalignment between modalities. For example, radar provides velocity for dynamic objects, while RCS from radar is related to objects' material, structure, and size. Integrating these radar-specific properties with the strong shape-aware LiDAR features can enhance the overall detection performance, especially for cars, which raise the most concern due to the special moving patterns and larger sizes.

Therefore, we propose a 4D radar-LiDAR fusion detection framework, MutualForce, that considers the advantages of both sensors and mutually enhances their own representations.
First, velocity and RCS information from the radar are chosen as the indicative features, guiding both radar and LiDAR geometric feature extractions. Successively, a shape-awareness network at the BEV level enhances the radar's BEV features using the shape information from LiDAR. The MutualForce outperforms existing methods on the View-of-Delft (VoD) dataset \cite{palffy2022multi}.
Overall, the main contributions are as follows:
\begin{itemize}
\item An Indicative Radar-Driven Bidirectional module (IRB), consisting of two branches, utilizes radar indicative information to guide geometric feature extraction of both radar and LiDAR point clouds. 
\item A multi-level Shape Awareness LiDAR-Driven Contrastive module (SALC) is introduced to enrich radar BEV features with LiDAR geometric information.
\item Extensive experiments on the VoD dataset \cite{palffy2022multi} demonstrate the effectiveness of our proposed method.
\end{itemize}

\section{Proposed Method}

\label{sec:2_Proposed Method}
\subsection{Overall Structure}

\begin{figure*}[t]
    \centering
    \includegraphics[width=0.9\textwidth,height=5.5cm]{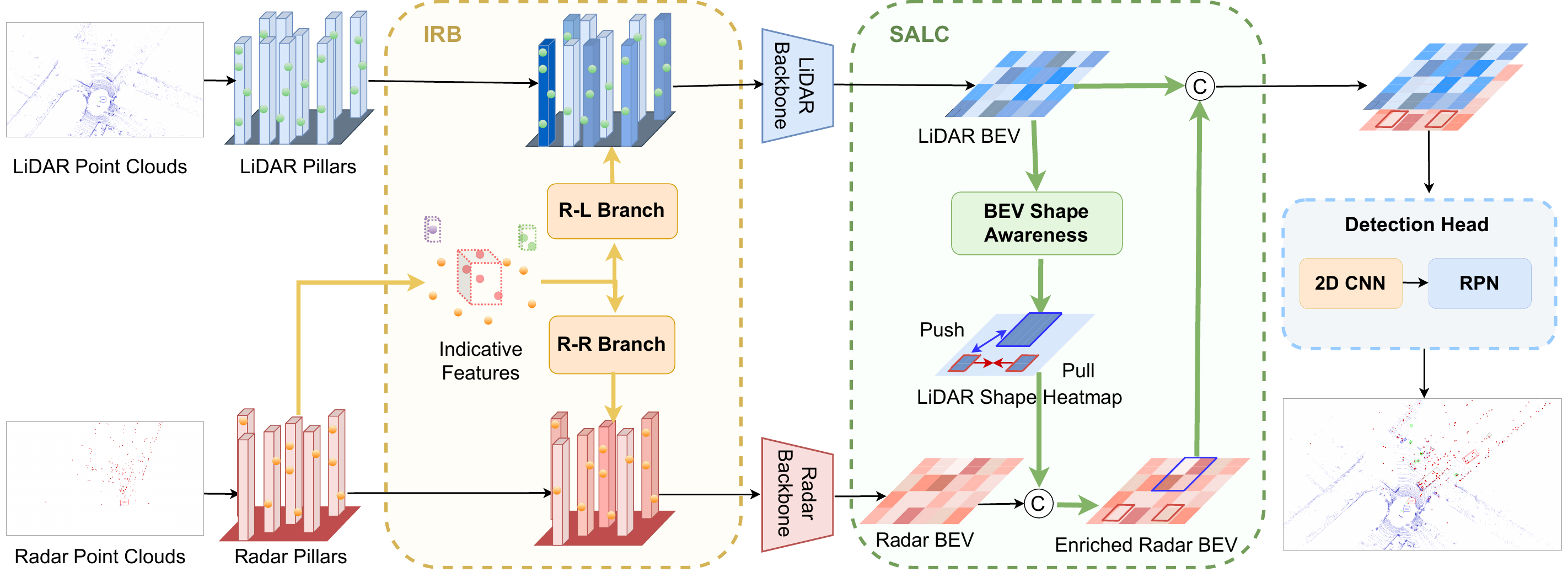} %
    \vspace{-2mm}
    \caption{The overall structure of MutualForce.}
    \label{fig: overall structure}
    \vspace{-4mm}
\end{figure*}
In this section, we introduce a fusion framework, MutualForce, which leverages the advantages of both sensors to mutually enhance their features. Its overall structure is illustrated in Fig. \ref{fig: overall structure}.
Initially, radar and LiDAR point clouds are mapped into pillars. Later, in the IRB module, the radar indicative features guide both radar and LiDAR geometric feature learning through a bidirectional attention mechanism. Subsequently, the multi-level shape information from LiDAR enriches radar BEV features by the SALC module. The final radar and LiDAR BEV features are concatenated and fed into the detection head to generate 3D proposals.

\subsection{Indicative Radar-Driven Bidirectional (IRB) Fusion}
\label{sec: IRB}
\begin{figure}[h]
    \vspace{-2mm}
    \centering
    \includegraphics[width=0.45\textwidth]{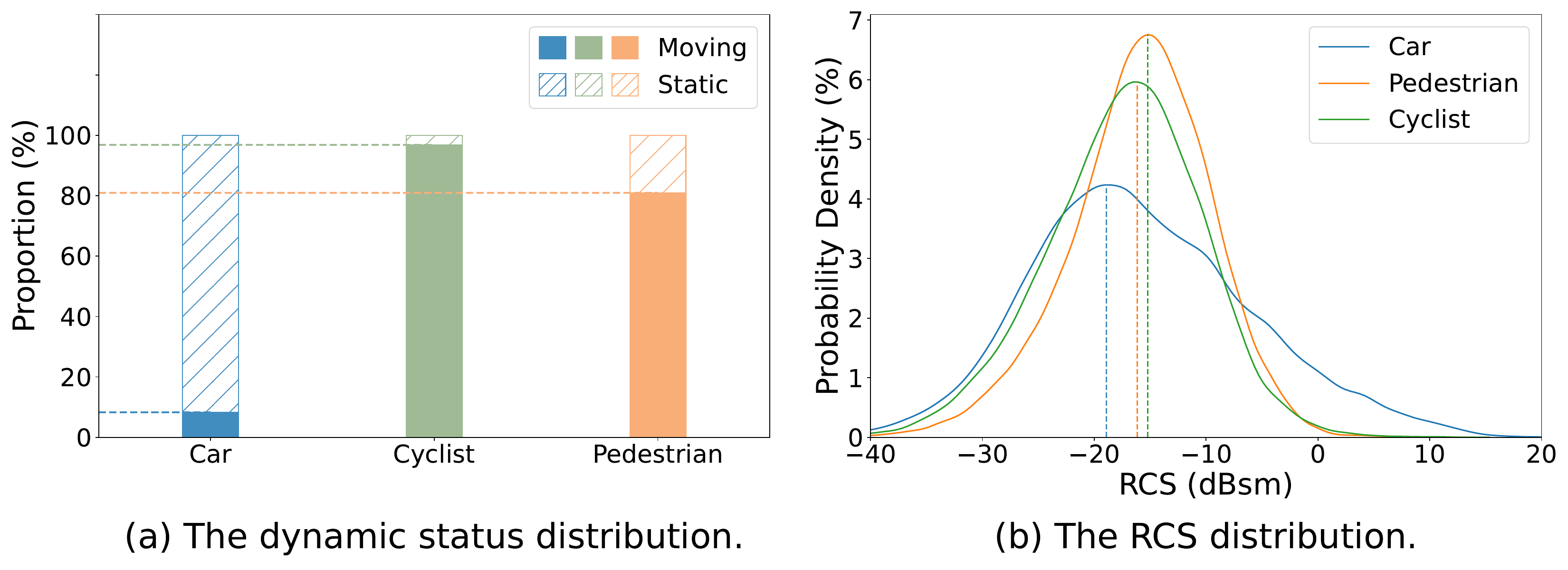} \vspace{-1mm}
    \caption{The distribution of radar indicative features for different road users in the VoD dataset.}
    \label{fig: v}
    \vspace{-0mm}
\end{figure}

Compared to high-speed vehicles, cyclists and pedestrians often have lower speed ranges with varied movement patterns. To effectively distinguish different moving road users, utilizing both the radar relative radial velocity and absolute radial velocity after ego-motion compensation becomes necessary.
However, in densely populated environments, there are more low-speed or stationary vehicles, as illustrated in Fig. \ref{fig: v}(a). They may have similar velocities as cyclists or pedestrians. In such cases, RCS from radar provides supplementary information since it relates to the object's structure, material, and size \cite{pang2024rcs}. As shown in Fig. \ref{fig: v}(b), cars have metallic surfaces with distinct reflections compared to pedestrians and cyclists. 

\begin{figure}[h]
    \centering
    \includegraphics[width=0.87\linewidth]{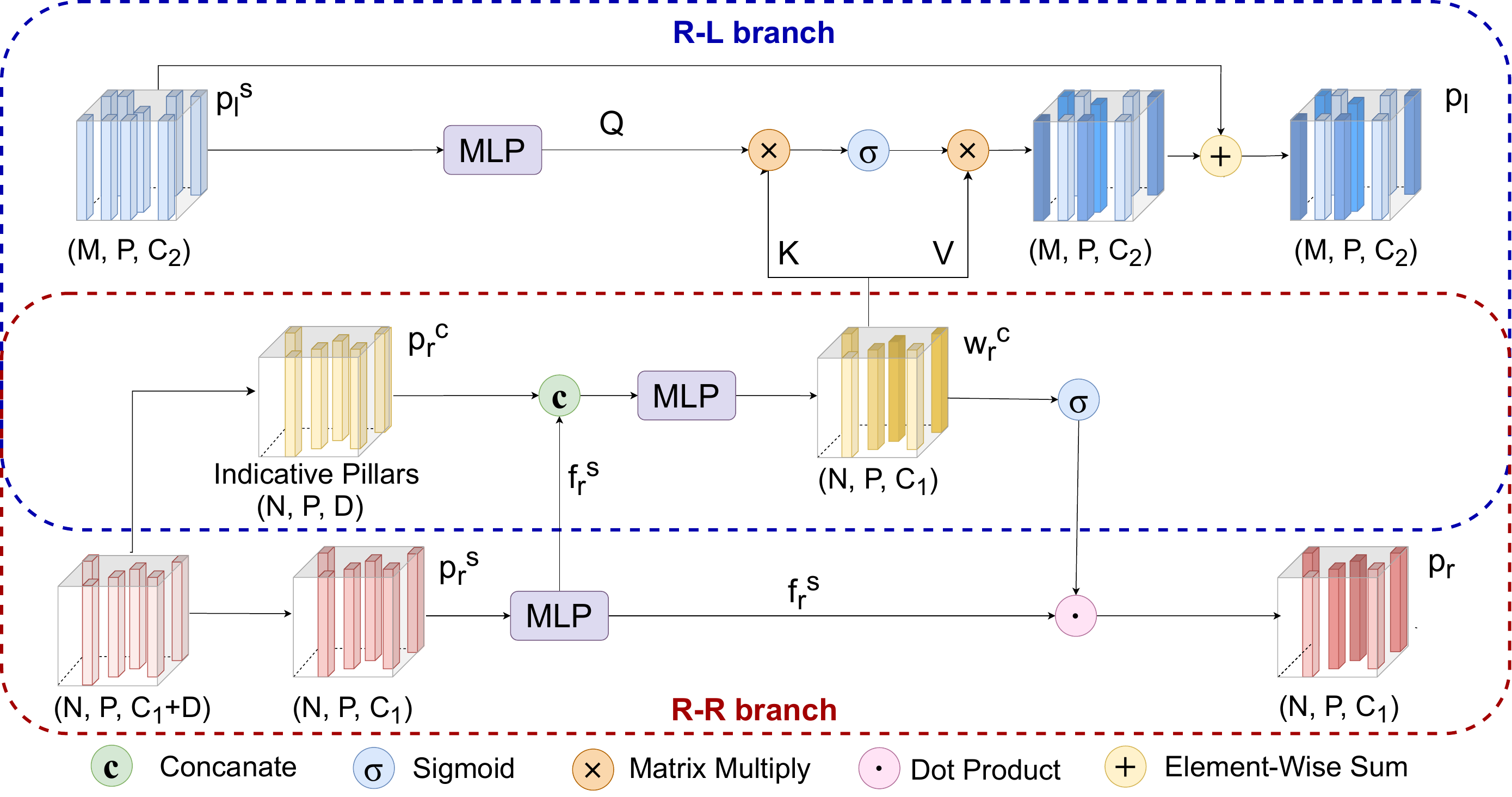}
    \vspace{-1mm}
    \caption{The Structure of IRB module.}
    \label{fig: the first idea}
    \vspace{-2mm}
\end{figure}

The structure of the IRB module is illustrated in Fig \ref{fig: the first idea}. It consists of two branches, R-R and R-L, guiding radar and LiDAR geometric feature learning. 
\vspace{-2mm}
After the pillarization of the raw point clouds, radar indicative pillars \( p_{r}^{c} \) and spatial pillars  \( p_{r}^{s} \) have the dimension of \( (N, P, D) \) and \( (N, P, C_1) \) while LiDAR pillars \( p_{l}^{s} \) are in dimension \( (M, P, C_2) \). \( M \) and \( N \) represent the number of pillars, \( P \) denotes the number of points in each pillar. \( C_1 \) and \( C_2 \) are the feature channels. \( D = 3\) indicates the three chosen radar indicative features, relative and absolute radial velocity, and RCS.

In the R-R branch, the radar intermediate geometric feature \( f_{r}^{s} \) is first captured from \( p_{r}^{s} \) through a Multilayer Perceptron (MLP).
Then the indicative feature \( p_{r}^{c} \) is concatenated with \( f_{r}^{s} \) to balance the indicative and geometric information and generate indicative weight \( w_{r}^{c} \). Subsequently, the indicative weight guides the radar spatial feature through a Sigmoid function and dot-product operation. The process of generating the final radar feature \( p_{r}^{} \) can be formulated as:
\vspace{-1mm}
\begin{equation}
w_{r}^{c} = \text{MLP}(p_{r}^{c} \mathbin{\text{\textcircled{c}}} f_{r}^{s}), \quad
p_{r} = \sigma(w_{r}^{c}) \cdot f_{r}^{s},
\vspace{-1mm}
\end{equation}
where \(\text{\textcircled{c}}\) denotes the concanation.
In the end, the radar representation is enhanced by R-R branch.

In the R-L branch, the final LiDAR feature \( p_{l}^{} \) is generated by a cross-attention mechanism. The LiDAR feature \( p_{l}^{s} \) undergoes an MLP as the query embedding \( Q \), while the indicative weight \( w_{r}^{c} \) serves as key \( K \) and value \( V \), bringing radar indicative information to LiDAR pillars
\vspace{-2mm}
\begin{equation}
p_{l} = p_{l}^{s} + \left( \frac{QK^T}{\sqrt{d_k}} \right) \cdot V,
\label{eq: pl}
\vspace{-0mm}
\end{equation}
where $ \left( \frac{QK^T}{\sqrt{d_k}} \right)$ represents the dot-product attention, \(d_k\) is key vectors' dimension. 
The dense LiDAR points can lead to redundant exploration of the background. By introducing radar indicative features, we include the dynamic and material information into LiDAR pillars. Therefore, LiDAR pillars can better identify important regions and foreground objects.

\subsection{Shape Awareness LiDAR-Driven Contrastive (SALC) Fusion}

\begin{figure}[h]
    \vspace{-2mm}
    \centering
    \includegraphics[width=0.91\linewidth]{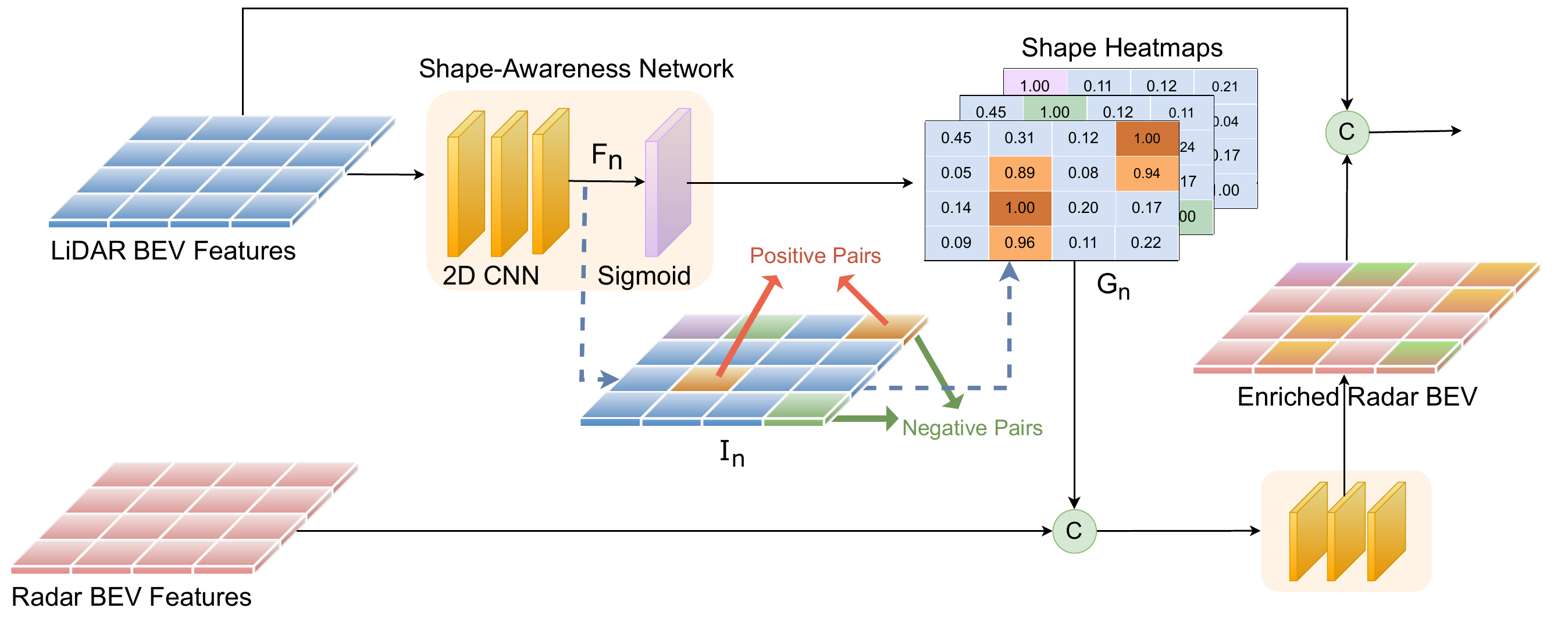}
    \vspace{-1mm}
    \caption{The structure of SALC module.} 
    \label{fig: the second idea}
    \vspace{-2mm}
\end{figure}

Radar points reflected from foreground objects often show extremely incomplete shapes \cite{peng2024transloc4d}. In contrast, LiDAR provides clearer fine-grained observation. It is particularly beneficial for cars with relatively symmetric shapes. Therefore, we propose the SALC module to enrich radar BEV features through the scene-level and instance-level shape information learned from LiDAR BEV features, as shown in Fig \ref{fig: the second idea}.  

In the SALC module, LiDAR BEV features \(B\in \mathbb{R}^{H \times W\times C} \) from PointPillars backbones \cite{lang2019pointpillars} are fed into a shape-awareness network composed of three 2D CNNs and one sigmoid layer. To obtain the global shape distribution, the shape-awareness network generates score-based shape heatmaps \( G_n \in \mathbb{R}^{H \times W} \), \( n = 1, \ldots, N \), where N is the number of classes. Each grid in \( G_n \) indicates the likelihood of belonging to the shape of $n$-th class. A threshold \( \tau \) is set to filter out the background grids. The remaining grids show the contours of the global scene. The awareness of scene-level shape distribution is supervised by the standard Focal Loss \(\mathcal{L}_{\text{cls}}\) \cite{lin2017focal}. 

Besides the scene-level shape distribution, 
instance-level understanding helps identify objects from different classes that show similar shapes due to occlusion and overlapping \cite{shen2023bsh}. An efficient way of pushing away close objects from different classes is contrastive learning. 
Assuming $F_n = \{ f(h, w) \mid h = 1, \ldots, H, w = 1, \ldots, W\}$ is the outputs before the sigmoid layer for the $n$-th class. According to CenterPoints \cite{yin2021center}, the grids in $G_n$ with value 1 indicate the objects' centers. Therefore, N instance indicators $I_n$ can be obtained by \(I_n = \{ f(h, w) \mid G_n(h, w) = 1 \}\), \( n = 1, \ldots, N \). 
Leveraging the centers of instances, a "push and pull" strategy from Multi-class Contrastive (MCcont) Loss \cite{xia2023coin} is employed to separate close shapes from different classes. 
First, a matrix \( \mathbf{S}^{N \times M} \) is defined from \( I_n \), where N denotes the number of classes, M is the maximum number of instance centers among all \( I_n \). The elements in each row of $S$ are the values in each \( I_n \), indicating different instances from the same class. Columns of $S$ are padded by randomly repeating the instances. \( \mathbf{S'} \) is later obtained by swapping the columns of \( \mathbf{S} \). Elements in the same row of \( \mathbf{S} \) and \( \mathbf{S'} \) are positive pairs from the same class, while elements in different rows are negative pairs. The instance-level contrastive shape separation is implemented as follows:
\vspace{-3mm}
\begin{equation}
\mathcal{L}_{\text{MCcont}} = -\frac{1}{N} \sum_{h=1}^{N}
\log \frac{\exp\left( \frac{d({S}(h,:), {S}'(:,h))}{M^2} \right)}{\sum_{w \ne h} \exp\left( \frac{d({S}(h,:), 
{S}'(:,w))}{M^2} \right)},
\label{eq:mc_loss}
\vspace{-1mm}
\end{equation}
where \( d( \cdot, \cdot ) \) denotes the distance calculated by element-wise product and sum.
The shape-aware loss \(\mathcal{L}_{\text{shape}}\) is the sum of Focal Loss \(\mathcal{L}_{\text{cls}}\) \cite{lin2017focal} for foreground shape distribution and MCcont Loss \(\mathcal{L}_{\text{MCcont}}\) \cite{xia2023coin} for close instances separation:
\vspace{-1mm}
\begin{equation}
\mathcal{L}_{\text{shape}} = \mathcal{L}_{\text{cls}} + \mathcal{L}_{\text{MCcont}}.
\vspace{-1mm}
\label{eq: shape}
\end{equation}
The generated shape heatmaps containing both global outline and instance details are integrated with radar BEV features via concatenation and convolution. Therefore, radar BEV features learn the multi-level shape information from LiDAR, supplementing missing and incomplete foreground geometric information. The enriched radar BEV features are concatenated with LiDAR BEV features to generate final proposals.
Our network is trained by a standard Region Proposal Network (RPN) detection loss with the shape-aware loss,
\vspace{-1mm}
\begin{equation}
\mathcal{L}_{\text{final}} = \mathcal{L}_{\text{RPN}} + \alpha \mathcal{L}_{\text{shape}},
\vspace{-1mm}
\end{equation}
Here, \(\alpha\) is set to 1.0.

\section{Experiments}
\label{sec:3_Experiments}
In this section, we compared MutualForce with other 3D detection algorithms. The models were trained with 2 NVIDIA Tesla A30 GPUs for 80 epochs with a batch size of 8. The implementation was built upon OpenPCDet library \cite{od2020openpcdet}.

\subsection{Dataset and Metrics}

Our approach is evaluated on the VoD dataset \cite{palffy2022multi}, containing 8600 frames of camera, LiDAR, and 4D radar data. Due to the unavailability of the test server, our evaluation is conducted on the validation set.
We utilize the Average Precision (AP) for each class and the mean Average Precision (mAP) across all classes to evaluate the results. An IoU threshold of 50\% is set for cars and 25\% for cyclists and pedestrians. The performance is assessed in the entire area and the driving corridor.

\subsection{Main Results}
\begin{figure}[t]
    \centering
    \vspace{-3mm}
    \begin{minipage}{0.005\linewidth} 
        \rotatebox[origin=l]{90}{\parbox{2cm}{\centering \scriptsize Pointpillars}}
    \end{minipage}%
    \begin{minipage}{0.995\linewidth}  
        \centering
        \begin{subfigure}{0.31\linewidth}
            \includegraphics[width=\linewidth]{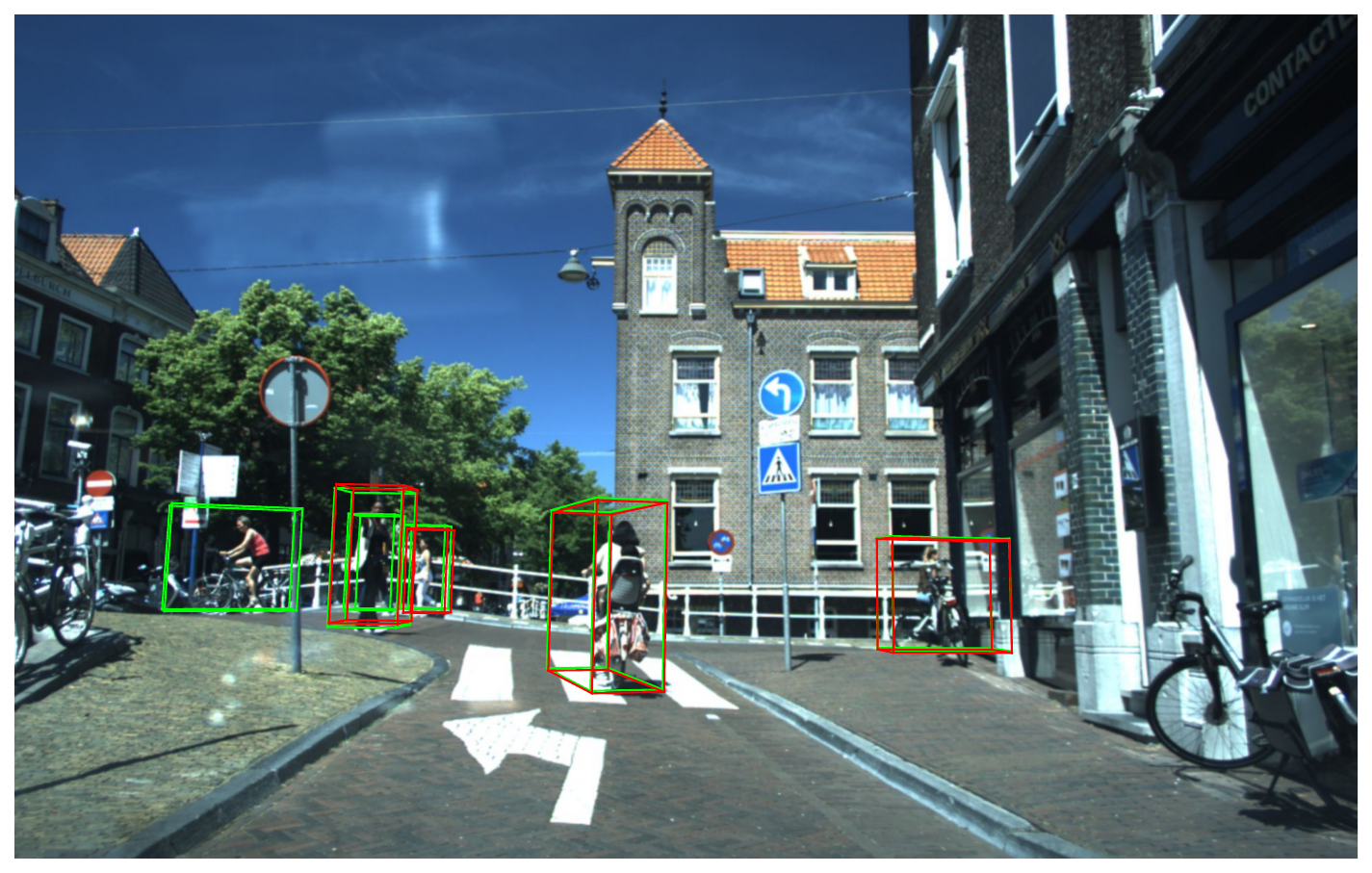}\hspace{-1mm}
        \end{subfigure}%
        \begin{subfigure}{0.31\linewidth}
            \includegraphics[width=\linewidth]{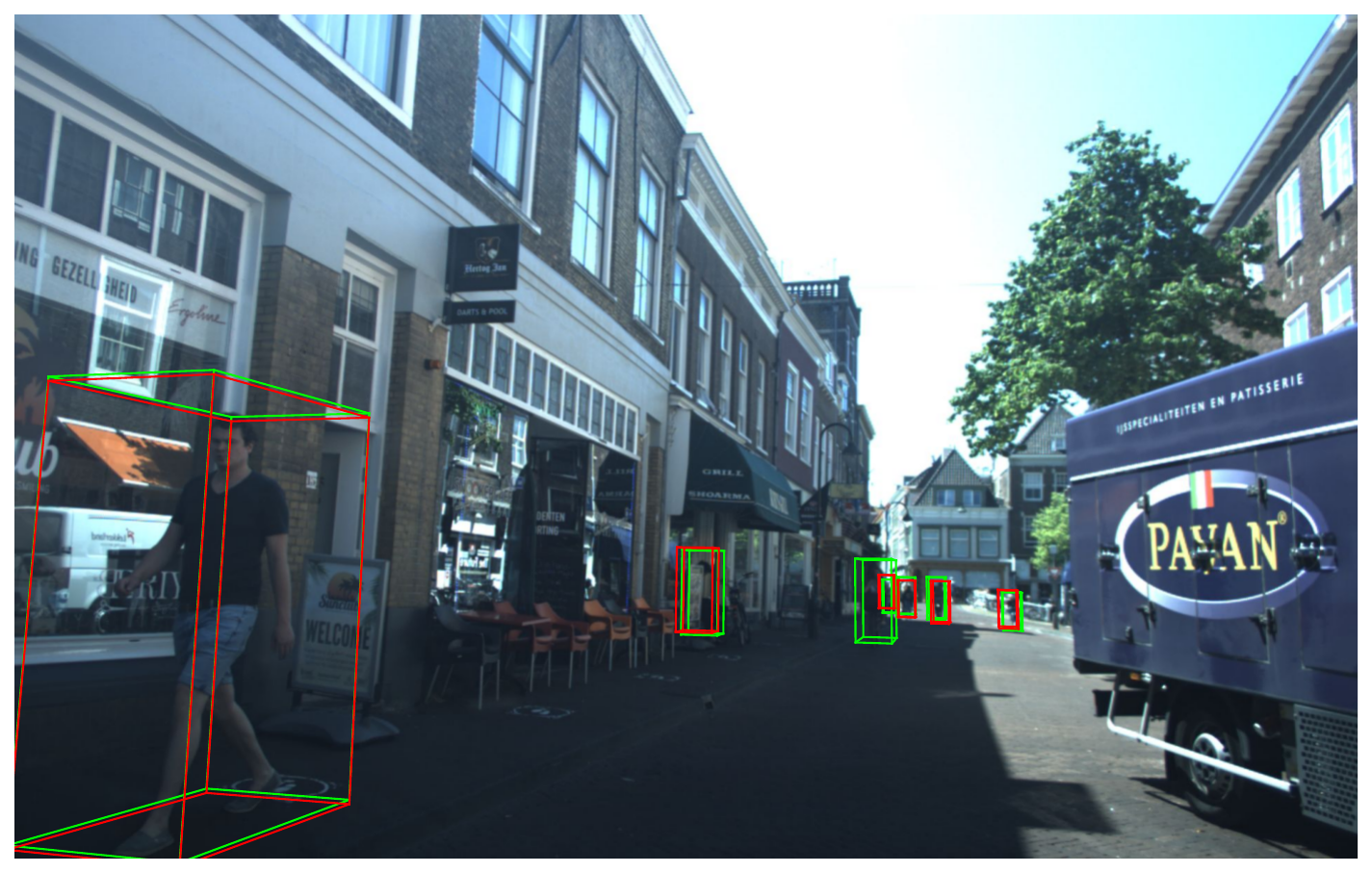}\hspace{-1mm}
        \end{subfigure}%
        \begin{subfigure}{0.31\linewidth}
            \includegraphics[width=\linewidth]{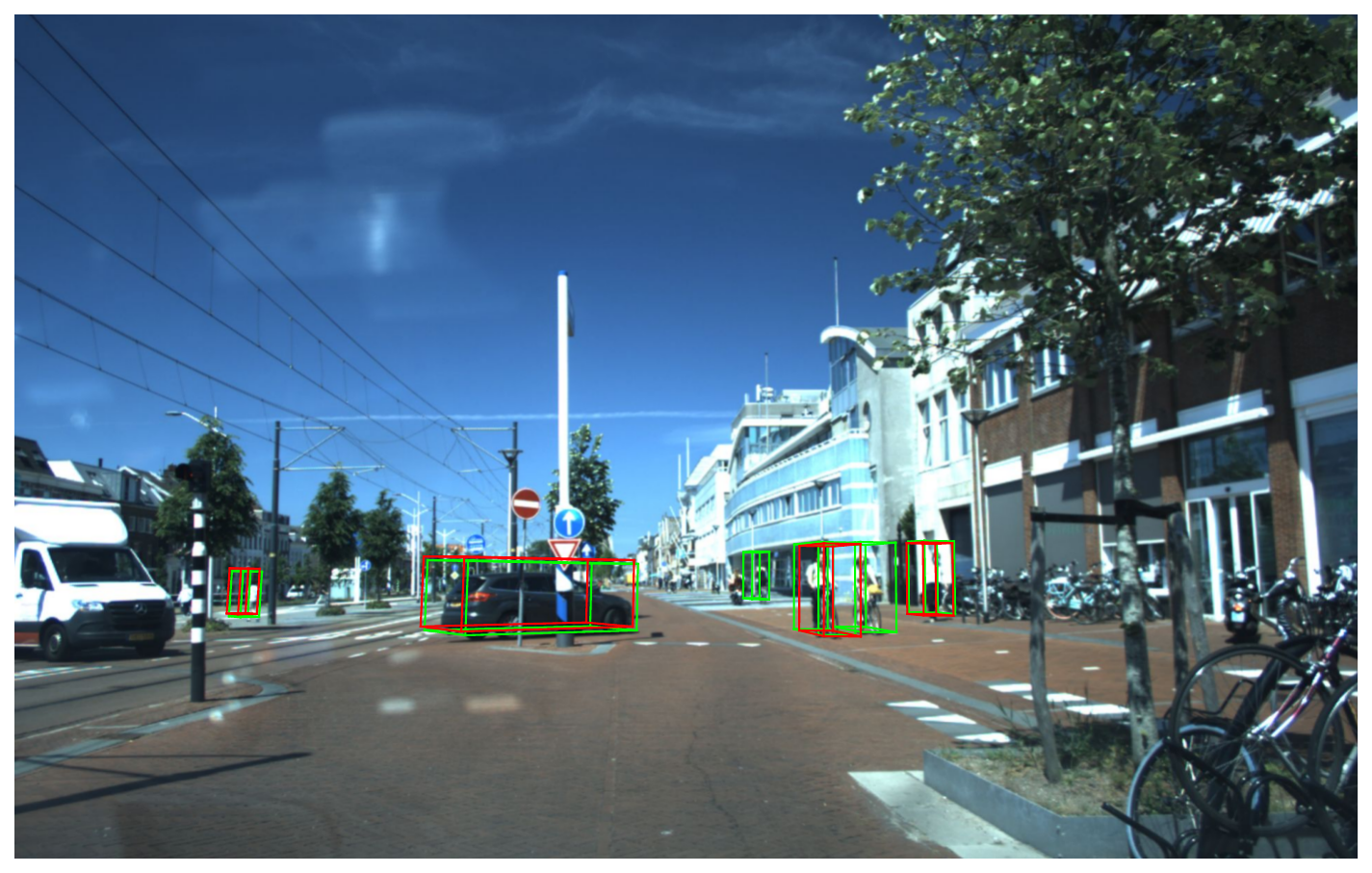}
        \end{subfigure}
    \end{minipage}
    \vspace{-4mm}

    \begin{minipage}{0.005\linewidth} 
        \rotatebox[origin=l]{90}{\parbox{2cm}{\centering \scriptsize Interfusion}}
    \end{minipage}%
    \begin{minipage}{0.995\linewidth}  
        \centering
        \begin{subfigure}{0.31\linewidth}
            \includegraphics[width=\linewidth]{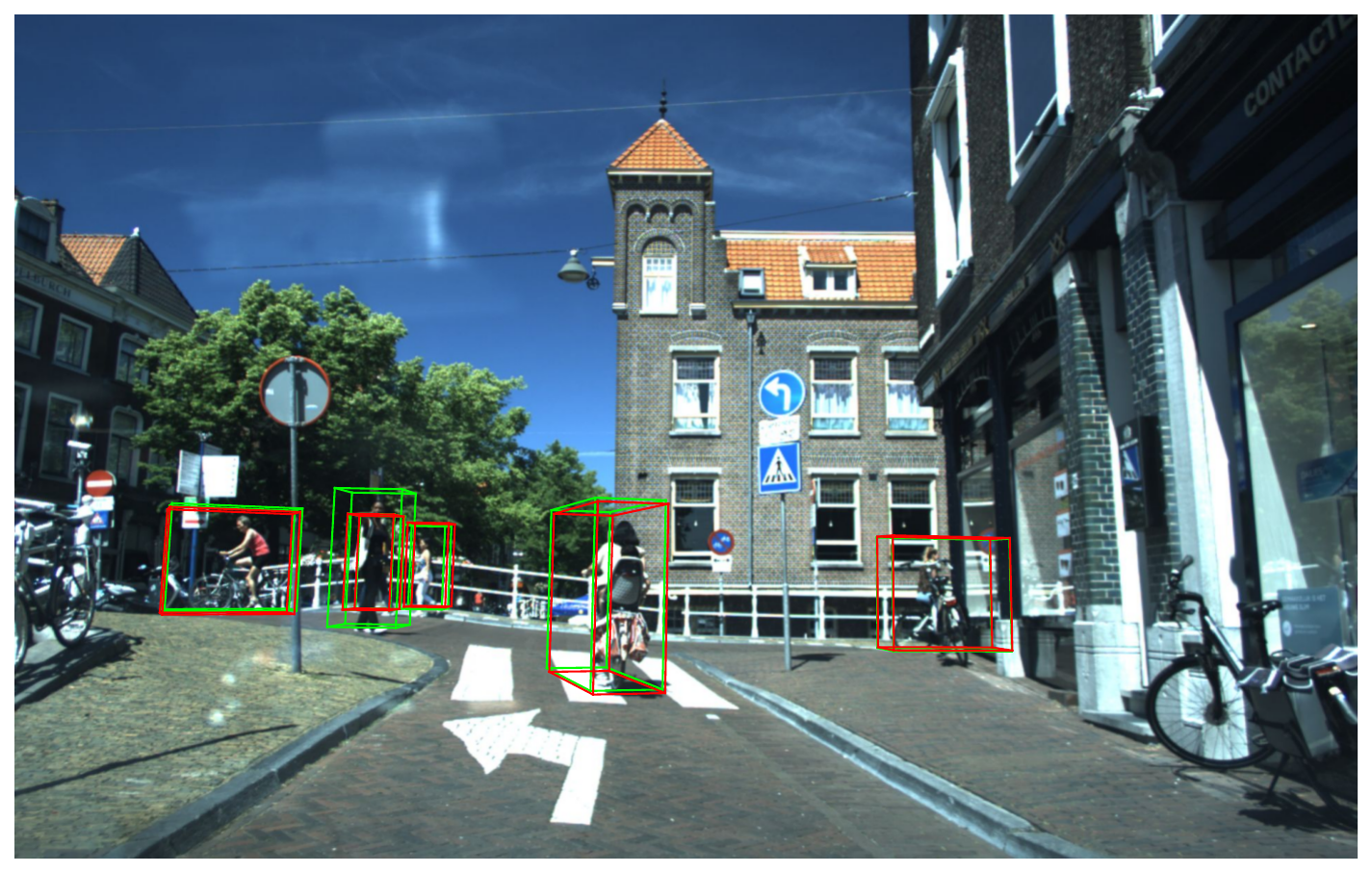}\hspace{-1mm}
        \end{subfigure}%
        \begin{subfigure}{0.31\linewidth}
            \includegraphics[width=\linewidth]{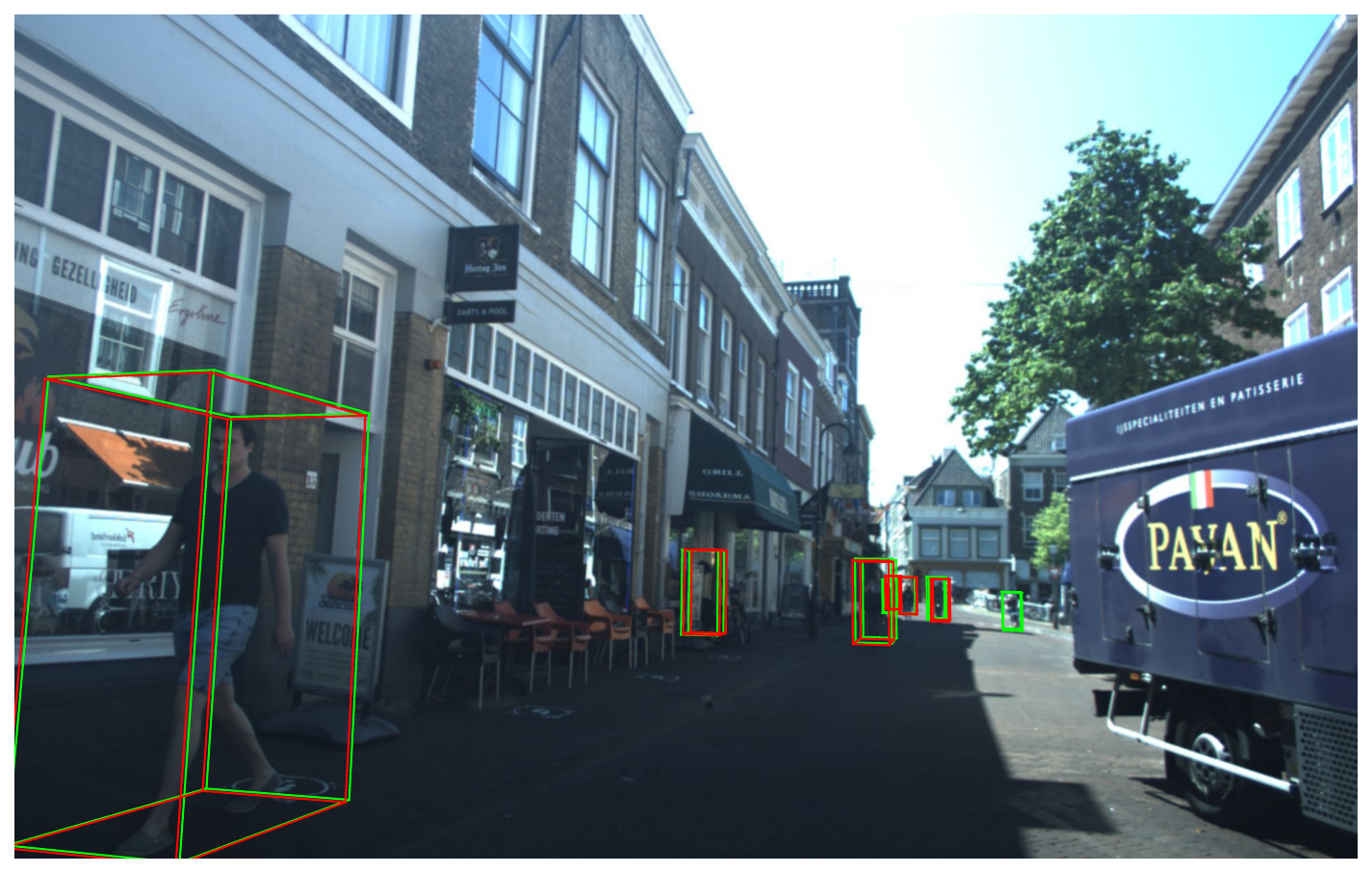}\hspace{-1mm}
        \end{subfigure}%
        \begin{subfigure}{0.31\linewidth}
            \includegraphics[width=\linewidth]{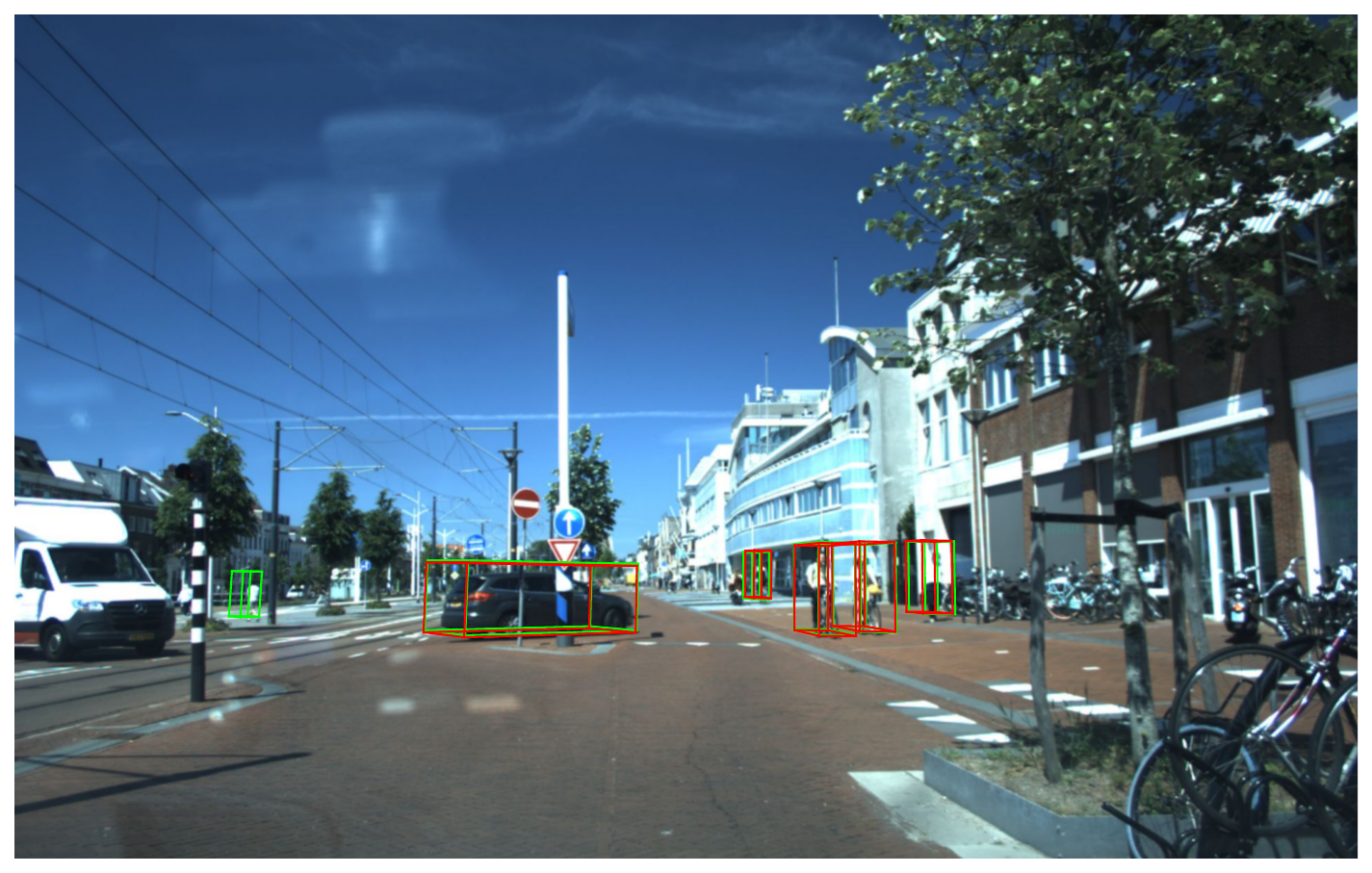}
        \end{subfigure}
    \end{minipage}
    \vspace{-4mm}
    
    \begin{minipage}{0.005\linewidth} 
        \rotatebox[origin=l]{90}{\parbox{2cm}{\centering \scriptsize Ours}}
    \end{minipage}%
    \begin{minipage}{0.995\linewidth}  
        \centering
        \begin{subfigure}{0.31\linewidth}
            \includegraphics[width=\linewidth]{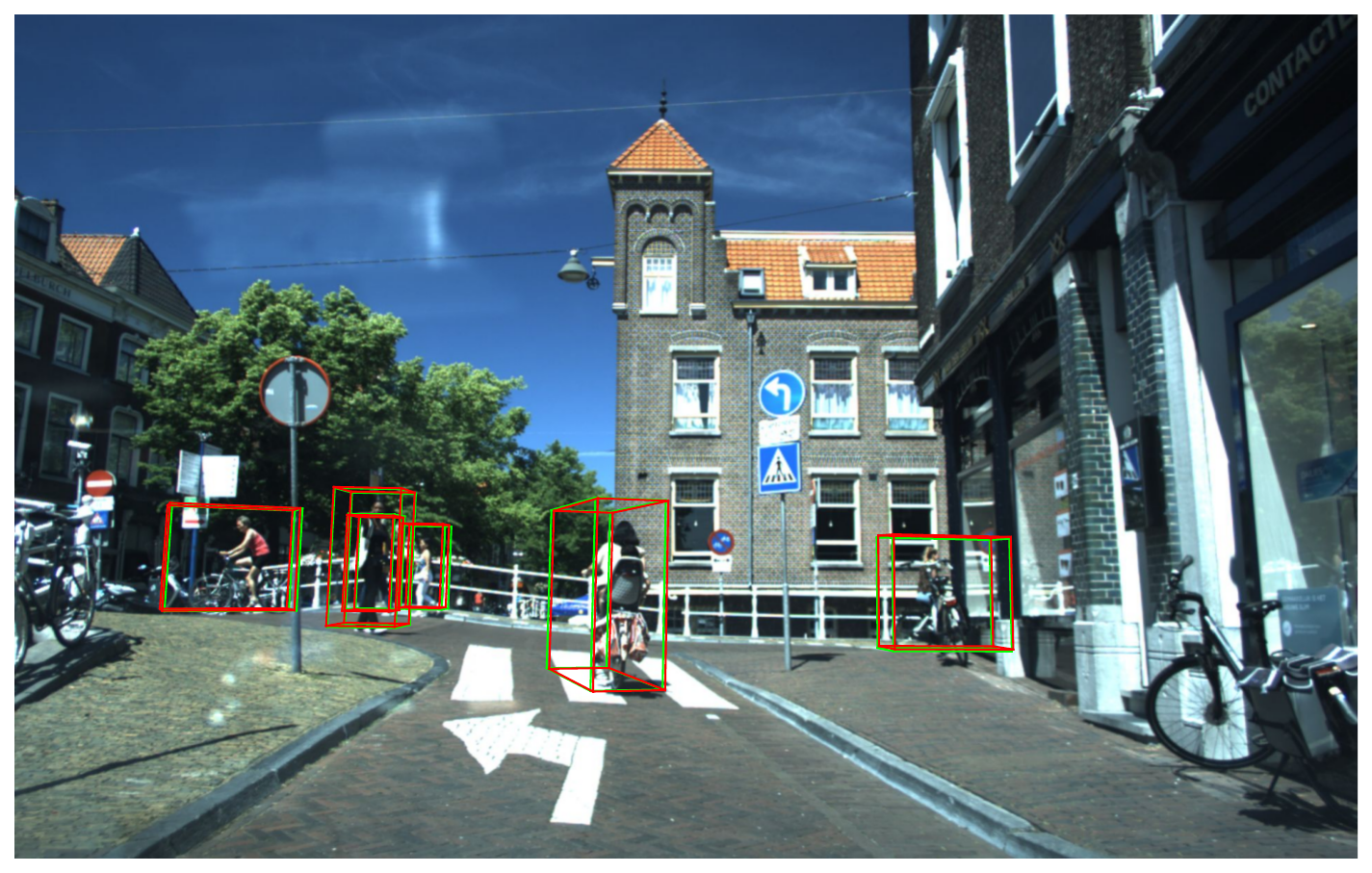}\hspace{-1mm}
        \end{subfigure}%
        \begin{subfigure}{0.31\linewidth}
            \includegraphics[width=\linewidth]{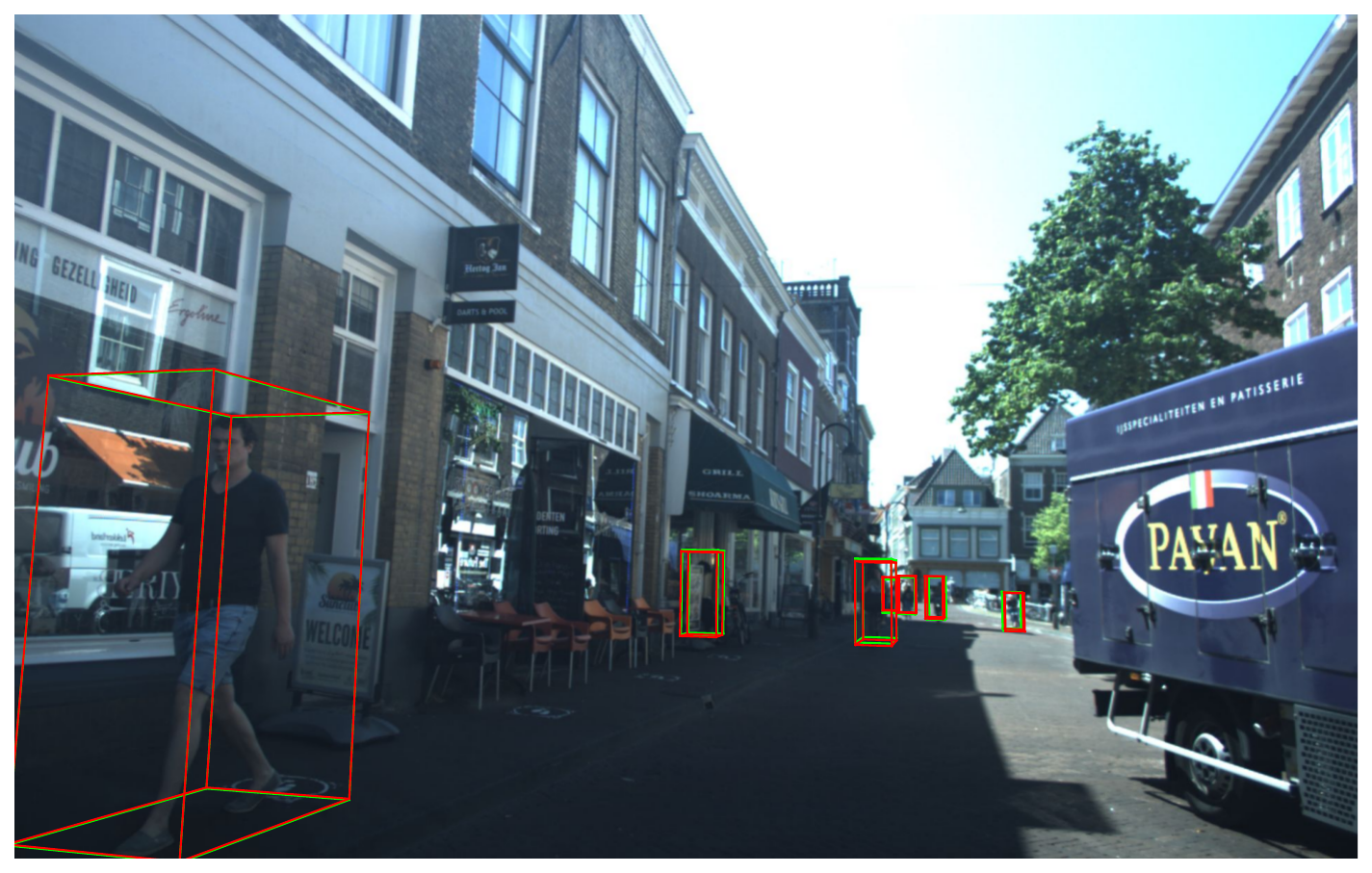}\hspace{-1mm}
        \end{subfigure}%
        \begin{subfigure}{0.31\linewidth}
            \includegraphics[width=\linewidth]{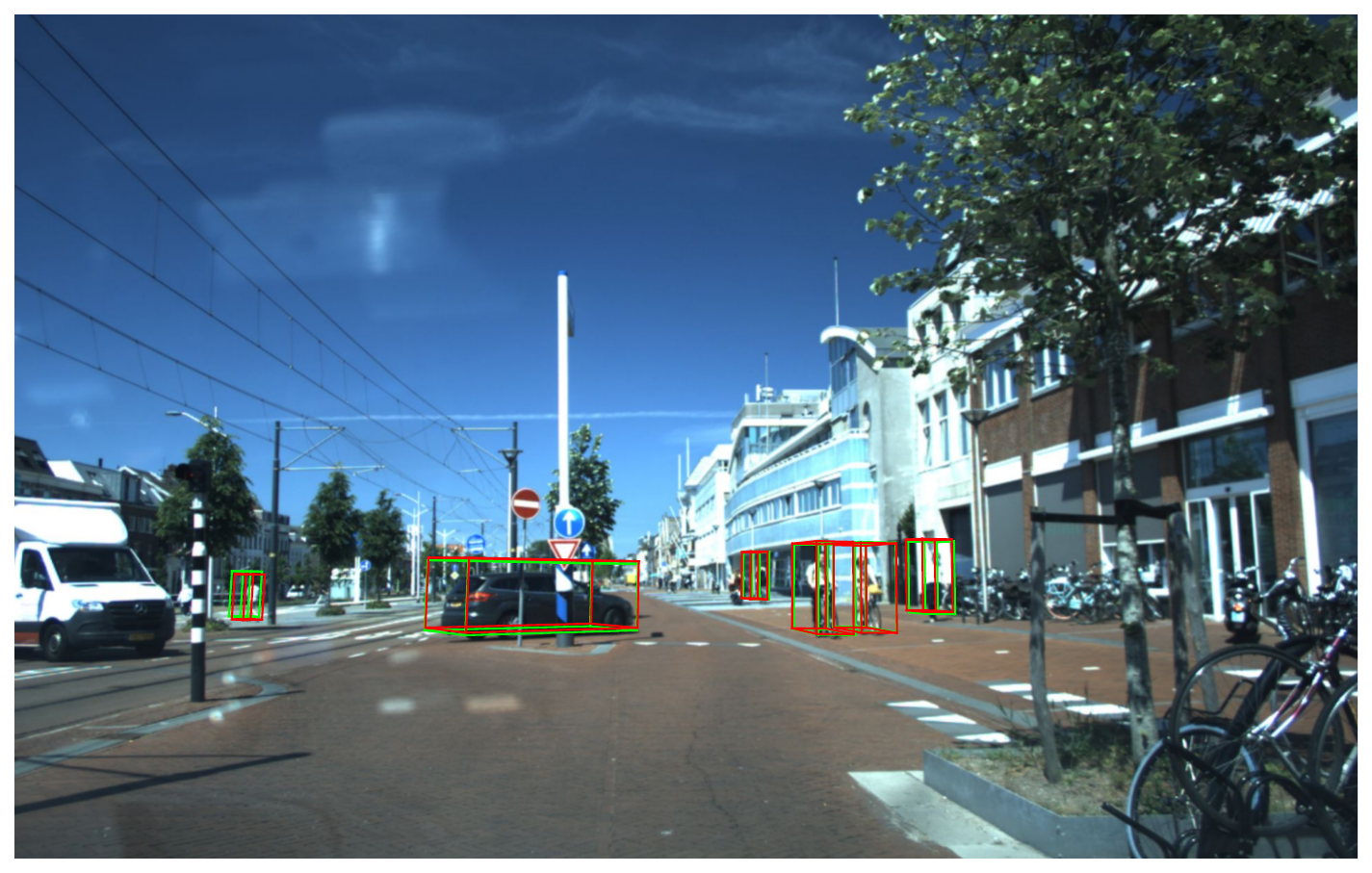}
        \end{subfigure}
    \end{minipage}
    \vspace{-2mm}
    \vspace{-1mm}
    \caption{Visualization of the detection on three methods. Ground truths are marked in green, while predicted boxes are red.}
    \label{fig: visualization}
    \vspace{-1mm}
\end{figure}

We performed extensive comparisons between MutualForce and existing single and multi-modal detection methods. The results are in Tables \ref{tab: main result vod}.
\setlength\tabcolsep{14pt}
\begin{table*}[ht]
\centering
\caption{Comparative AP results on VoD val. set. The values are in \%. The best results are bold.}
\label{tab: main result vod}
\vspace{-2mm}
\begin{tabular}{c|c|cccc|cccc}
\hline
\multirow{2}{*}{Methods} &\multirow{2}{*}{Modality} & \multicolumn{4}{c|}{All area}   & \multicolumn{4}{c}{Driving Corridor}\\
\cline{3-10}
 &  &Car & Ped. &Cyc. & mAP & Car & Ped. & Cyc. &  mAP\\
\hline
MVFAN\textsuperscript{\dag} \cite{yan2023mvfan} &R &38.12 &30.96  &66.17  &45.08  &71.45 &40.21 &86.63 &66.10 \\
PV-RCNN\textsuperscript{\dag} \cite{shi2020pv} &R &41.65   &38.82  &58.36  &46.28  &72.00  &43.53  &78.32  &64.62 \\
SMURF \cite{10274127} &R &42.31 &39.09  &71.50  &50.97  &71.74  &50.54  &86.87  &69.72 \\
MUFASA\textsuperscript{\dag} \cite{peng2024mufasamultiviewfusionadaptation} &R &43.10  &38.97 &68.65 &50.24 &72.50  &50.28 &88.51  &70.43\\
\hdashline
BEVFusion \cite{10160968}  &R+C &37.85 &40.96 &68.95 &49.25 &70.21 &45.86  &89.48  &68.52 \\
RCFusion \cite{zheng2023rcfusion}  &R+C &41.70 &38.95 &68.31 &49.65 &71.87 &47.50  &88.33  &69.23 \\
RCBEVDet \cite{lin2024rcbevdet} &R+C &40.63 &38.86  &70.48  &49.99  &72.48  &49.89  &87.01  &69.80\\
LXL \cite{xiong2023lxl} &R+C &42.33 &49.48 &77.12 &56.31 &72.18 &58.30 &88.31 &72.93 \\
\hdashline
Pointpillars\textsuperscript{\dag} \cite{lang2019pointpillars} &L &65.55  &55.71  &72.96  &64.74  &81.10  &67.92  &88.96  &79.33 \\
LXL-Pointpillars \cite{xiong2023lxl} &L &66.60 &56.10 &75.10 &65.90 &-  &-  &-  &- \\
InterFusion\textsuperscript{\dag} \cite{wang2022interfusion} &R+L &67.50 &63.21  &\textbf{78.79} &69.83 &88.11  &74.80  &87.50  &83.47\\
\hdashline
MutualForce (Ours) &R+L &\textbf{71.67} &\textbf{66.26} &77.35 &\textbf{71.76}  &\textbf{92.31} &\textbf{76.79} &\textbf{89.97} &\textbf{86.36}  \\
\hline
\end{tabular}
\\[1pt]
\scriptsize{R, C, and L denote the radar, camera, and LiDAR. \textsuperscript{\dag} indicates our reproduced results.}
\vspace{-3mm}
\label{tab: main result vod}
\end{table*}
From Table \ref{tab: main result vod}, radar-only and radar-camera methods yield unsatisfactory detection due to the sparsity of radar point clouds. In contrast, incorporating LiDAR significantly enhances the performance. Notably, our method outperforms other approaches with the mAP of 71.76\% across the entire area and 86.36\% in the driving corridor.
Although most cars are static in the VoD dataset, as in Fig. \ref{fig: v}(a), with the radar indicative information and LiDAR shape awareness, our method achieves an improvement of 4.17\% and 4.20\% AP for cars in two regions compared to Interfusion \cite{wang2022interfusion}. For small objects, our approach delivers better results, particularly in driving corridors, with a 1.99\% AP increase for pedestrians and 2.47\% for cyclists.
Fig. \ref{fig: visualization} visualize the detection results of Pointpillars \cite{lang2019pointpillars}, Interfusion \cite{wang2022interfusion}, and MutualForce. Our method performs better with fewer false negatives.

\subsection{Ablation Study}

\textbf{Analysis of different modules:}
Experiments with different modules are presented in Table \ref{tab: modules}.
\setlength\tabcolsep{2.7pt}
\begin{table}
\centering
\caption{Analysis of each module. The values are in \%.}
\vspace{-2mm}
\label{tab: modules}
\begin{tabular}{cc|c|cccc|cccc}
\hline
\multicolumn{2}{c|}{IRB} & \multirow{2}{*}{SALC} & \multicolumn{4}{c|}{All area} & \multicolumn{4}{c}{Driving Corridor} \\
\cline{1-2} \cline{4-11}
 R-R & \multicolumn{1}{c|}{R-L} & & Car & Ped. & Cyc. & mAP & Car & Ped. & Cyc. & mAP \\
\hline
&  & &61.14 &50.79 &67.32 &59.75 &81.99 &62.14 &86.40 &76.84 \\
$\checkmark$ &  & & 61.48 &57.16 &68.60 &62.41 &84.42 &70.67 &88.49 &81.19 \\
&$\checkmark$ & &69.64 &61.77 &77.76 &69.72 &90.87 &74.35 &89.57 &84.93 \\
$\checkmark$&$\checkmark$ & &67.75 &61.72 &\textbf{79.41} &69.63 &90.96 &73.99 &\textbf{90.21} &85.05\\
& &$\checkmark$ &69.53 &59.58 &74.77 &67.96 &90.74 &72.69 &88.81 &84.08 \\
$\checkmark$ &$\checkmark$ &$\checkmark$ &\textbf{71.67} &\textbf{66.26} &77.35 &\textbf{71.76} &\textbf{92.31} &\textbf{76.79} &89.97 &\textbf{86.36} \\
\hline
\end{tabular}
\vspace{-4mm}
\end{table}
We first remove the IRB and SALC modules and directly concatenate two BEV features after feature extraction as the baseline. As shown in Table \ref{tab: modules}, the R-L branch enhances the mAP by 9.97\% and 8.09\% in two regions, demonstrating that radar's indicative information effectively guides LiDAR's geometric features. Besides, the SALC module helps separate occluded objects by multi-level shape information, especially for cars and pedestrians, achieving an improvement of 8.39\% and 8.79\% in all areas.

\textbf{Analysis of different indicative features:}
The ablation studies on different indicative features are in Table \ref{tab: indicative feature}. The baseline is implemented by removing the IRB module from the overall structure.
\setlength\tabcolsep{3.5pt}
\begin{table}
\centering
\caption{Analysis of indicative features. The values are \%. }
\vspace{-2mm}
\label{tab: indicative feature}
\begin{tabular}{ccc|cccc|cccc}
\hline
\multicolumn{3}{c|}{IRB} & \multicolumn{4}{c|}{All area} & \multicolumn{4}{c}{Driving Corridor} \\
\cline{1-11}
 $V_r$& $V_a$ & RCS & Car & Ped. & Cyc. & mAP & Car & Ped. & Cyc. & mAP \\
\hline
& & &69.53 &59.58 &74.77 &67.96 &90.74 &72.69 &88.81 &84.08 \\
$\checkmark$ &   & & 69.54 & 60.36 & 74.41 &68.10 & 91.04 & 72.98 & 88.66 &84.23 \\
&$\checkmark$ & & 69.68 & 62.23 & 77.56 &69.82 & 90.78 &75.27 & 90.05 &85.37 \\
 &  &$\checkmark$ &70.61 & 60.86 & 76.12 &69.20 & 91.27 & 74.30 &87.89 &84.49 \\
 $\checkmark$ &$\checkmark$  & & 69.23 &\textbf{67.64} & 77.29 &71.39 & 89.36 & 74.55 & 88.92 &84.28 \\
  &$\checkmark$ &$\checkmark$ & 69.98 & 61.53 &\textbf{78.76} &70.09 & 91.58 & 68.17 &\textbf{90.29} &83.35  \\
$\checkmark$ & &$\checkmark$ & 69.34 &67.08 & 77.14 &71.19 & 91.43 & 74.53 & 88.31 &84.76 \\
$\checkmark$&$\checkmark$ &$\checkmark$&\textbf{71.67} &66.26 &77.35 &\textbf{71.76} &\textbf{92.31} &\textbf{76.79} &89.97 &\textbf{86.36} \\
\hline
\end{tabular}
\\[1pt]
\scriptsize{$V_r$ denotes the relative radial velocity. $V_a$ denotes the absolute radial velocity.}
\vspace{-1mm}
\end{table}
According to Table \ref{tab: indicative feature}, velocity is important in detecting small objects. Leveraging relative and absolute radial velocity improves the AP by 8.06\% for pedestrians and 2.52\% for cyclists in the entire scene. 
Besides, RCS enhances the detection due to the different surfaces' materials. 

\textbf{Analysis of the BEV shape heatmaps in SALC:}
\( \tau \) is used in SALC to determine whether a grid belongs to a shape. Results with different \( \tau \) are in Table \ref{tab:threshold_effect}.
\setlength\tabcolsep{5pt}
\begin{table}
\centering
\caption{Analysis of BEV threshold. The values are in \%. }\label{tab:threshold_effect}
\vspace{-2mm}
\begin{tabular}{c|cccc|cccc}
\hline
\multirow{2}{*}{\( \tau \)}  & \multicolumn{4}{c|}{All area}   & \multicolumn{4}{c}{Driving Corridor}\\
\cline{2-9}
 &Car & Ped. & Cyc. & mAP & Car & Ped. & Cyc. & mAP\\
\hline
0.05 &69.74 &56.78 &\textbf{78.75} &68.42 &91.52 &74.16 &89.86 &85.18 \\
0.1 &\textbf{71.67} &\textbf{66.26} &77.35 &\textbf{71.76} &\textbf{92.31} &\textbf{76.79} &\textbf{89.97} &\textbf{86.36} \\
0.2 &68.73 &58.76 &76.84 &68.11 &90.88 &68.19 &87.18 &82.08 \\
\hline
\end{tabular}
\vspace{-4mm}
\end{table}
Our model achieves the best mAP when \( \tau \) equals 0.1, identifying the most shapes in LiDAR BEV features, especially for cars and pedestrians.

\section{Conclusion}
\label{sec:5_Conclusion} 

Our MutualForce fully utilizes the respective strengths of radar and LiDAR to mutually enhance their representations. In the pillar stage, the radar's indicative information guides the geometric feature extraction of both modalities. In the BEV stage, the multi-level shape information from LiDAR enriches the radar's BEV features. 
the . Besides, our model delivers a real-time performance at 14.45 FPS with 20.03M parameters.

\section*{Acknowledgment}

This research was conducted as part of the DELPHI project, funded by the European Union under grant agreement No. 101104263, and the SICHER project, supported by the German Federal Ministry for Economic Affairs and Climate Action through the funding program “Neue Fahrzeug- und Systemtechnologien”.
The views and opinions expressed do not necessarily reflect those of the European Union or the European Climate, Infrastructure, and Environment Executive Agency. Neither the European Union nor the granting authority can be held responsible for them.

\end{document}